\documentclass[runningheads]{llncs}
\usepackage[T1]{fontenc}

\usepackage{graphicx,verbatim}
\usepackage{amsmath}
\usepackage{amsfonts}
\usepackage{amssymb}
\usepackage{algorithm,algorithmic}
\usepackage{booktabs}
\usepackage{comment}
\begin{document}
\title{RAFM: Retrieval-Augmented Flow Matching for Unpaired CBCT-to-CT Translation}
\titlerunning{RAFM: Retrieval-Augmented Flow Matching}
%

\author{
Xianhao Zhou\inst{1} \and
Jianghao Wu\inst{2} \and
Lanfeng Zhong\inst{1,3} \and
Ku Zhao\inst{1} \and
Jinlong He\inst{1} \and
Shaoting~Zhang\inst{1,3} \and
Guotai Wang\inst{1,3}\thanks{Corresponding author: Guotai Wang (guotai.wang@uestc.edu.cn).}
}
\authorrunning{X. Zhou et al.}

\institute{
School of Mechanical and Electrical Engineering,\\
University of Electronic Science and Technology of China, Chengdu, China
\and
Faculty of Information Technology, \\
Monash University, Melbourne, VIC 3800, Australia
\and
Shanghai Artificial Intelligence Laboratory, Shanghai, China
}



  
\maketitle              
\begin{abstract}
Cone-beam CT (CBCT) is routinely acquired in radiotherapy but suffers from severe artifacts and unreliable Hounsfield Unit (HU) values, limiting its direct use for dose calculation. Synthetic CT (sCT) generation from CBCT is therefore an important task, yet paired CBCT--CT data are often unavailable or unreliable due to temporal gaps, anatomical variation, and registration errors. In this work, we introduce rectified flow (RF) into unpaired CBCT-to-CT translation in medical imaging. Although RF is theoretically compatible with unpaired learning through distribution-level coupling and deterministic transport, its practical effectiveness under small medical datasets and limited batch sizes remains underexplored. Direct application with random or batch-local pseudo pairing can produce unstable supervision due to semantically mismatched endpoint samples. To address this challenge, we propose Retrieval-Augmented Flow Matching (RAFM), which adapts RF to the medical setting by constructing retrieval-guided pseudo pairs using a frozen DINOv3 encoder and a global CT memory bank. This strategy improves empirical coupling quality and stabilizes unpaired flow-based training. Experiments on SynthRAD2023 under a strict subject-level true-unpaired protocol show that RAFM outperforms existing methods across FID, MAE, SSIM, PSNR, and SegScore. 
The code is available at https://github.com/HiLab-git/RAFM.git.
\end{abstract}
%
%
%
\section{Introduction}

Computed tomography (CT) is the standard imaging modality for radiotherapy treatment planning because its Hounsfield Unit (HU) values provide reliable electron-density information for dose calculation~\cite{sct_review}. In routine image-guided radiotherapy, cone-beam CT (CBCT) can be repeatedly acquired on treatment machines and is therefore valuable for adaptive radiotherapy workflows~\cite{cbct_application_review}. However, CBCT often suffers from severe artifacts and inaccurate HU values, which limits its direct use for dose computation~\cite{cbct_artifacts}. This has motivated CBCT-to-CT translation, i.e., synthesizing CT-like images from CBCT for downstream radiotherapy applications~\cite{sct_review}. In practice, this problem is often inherently unpaired: paired CBCT--CT data are difficult to curate and align accurately, and even nominally paired scans may be affected by temporal gaps, anatomical variation, and registration errors. Thus, unpaired CBCT-to-CT translation is a practically important setting for improving CT-domain realism while preserving patient anatomy~\cite{cbct2ct_cycleGAN}.

Existing unpaired CBCT-to-CT methods are mainly based on Generative Adversarial Networks (GANs) or diffusion/Schr\"odinger-bridge (SB) models. GAN-based methods, including cycle-consistent and contrastive variants~\cite{cbct2ct_cycleGAN,GcGAN,CuT}, can produce strong visual results but inherit adversarial-training instability and sensitivity to architecture and loss balancing. Diffusion/SB-based methods have also shown promising performance in unpaired medical image translation~\cite{syndiff,UNSB,ACSB}, but typically require more complex training pipelines and may still involve adversarial components. These limitations motivate a fully non-adversarial alternative for anatomy-sensitive unpaired translation, where stable optimization and reliable structure preservation are particularly important.

Flow Matching and Rectified Flow (RF) provide a fully non-adversarial direction for unpaired translation by modeling it as deterministic transport between distributions~\cite{flow_matching,rectified_flow}. In RF, the training objective is defined on endpoint couplings whose marginals match the source and target domains, so it does not require voxel-aligned paired correspondences in principle~\cite{rectified_flow}. Despite this appealing property, RF has not yet been systematically explored for medical image-to-image translation. A key practical challenge is that medical datasets are typically small, making finite-sample couplings a poor approximation of the desired distribution-level coupling; consequently, random or semantically arbitrary endpoint pairing can introduce noisy transport targets and harm anatomy-sensitive translation~\cite{FM_pairing}. To mitigate transport noise, prior work in natural-image settings has proposed constructing couplings within each mini-batch to favor more compatible endpoints~\cite{FM_batch_OT}. However, medical images are high-resolution and memory-intensive, so training often relies on small batch sizes, leaving batch-local coupling with an overly limited candidate pool and thus limited effectiveness.

To address these limitations, we introduce Retrieval-Augmented Flow Matching (RAFM), which enhances unpaired RF training through retrieval-guided coupling. RAFM retains the non-adversarial RF formulation while replacing random or batch-local pseudo pairing with global retrieval from a CT memory bank. Specifically, we use a frozen DINOv3~\cite{dinov3} encoder to construct a feature space for CT slices and retrieve a feature-similar CT slice for each CBCT slice to form pseudo endpoint pairs for flow matching. This retrieval-augmented strategy improves the empirical coupling quality under small-data, small-batch conditions while remaining strictly unpaired, as no subject identity or paired CBCT--CT correspondence is used. We evaluate RAFM on SynthRAD2023 under a strict subject-level true-unpaired protocol and demonstrate consistent improvements in image quality, distributional realism, and anatomy-related consistency.

\section{Method}

\begin{figure*}[t]
\centering
\includegraphics[width=\textwidth]{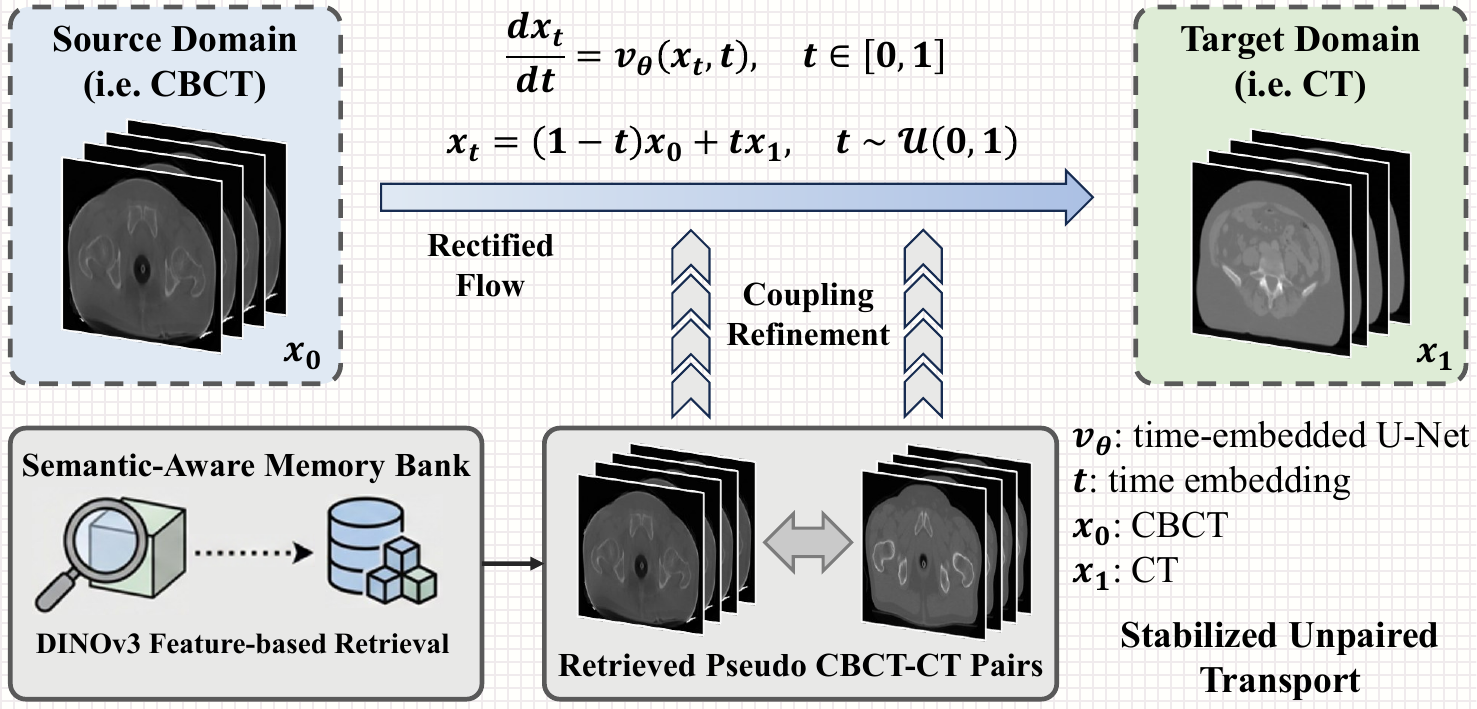}
\caption{Overview of Retrieval-Augmented Flow Matching (RAFM).
}
\label{fig:method}
\end{figure*}

\subsection{Problem Setup and Overview}


We consider unpaired CBCT-to-CT translation with two training sets,
$\mathcal{D}^{\text{tr}}_{\text{cbct}}=\{x_i^{\text{cbct}}\}_{i\in\mathcal{A}}$
and
$\mathcal{D}^{\text{tr}}_{\text{ct}}=\{x_i^{\text{ct}}\}_{i\in\mathcal{B}}$,
where $\mathcal{A}\cap\mathcal{B}=\varnothing$.
These two sets are sampled from the CBCT and CT marginals
$\pi_{\text{cbct}}$ and $\pi_{\text{ct}}$, respectively.
Given $x^{\text{cbct}}\sim\pi_{\text{cbct}}$, the goal is to synthesize a CT-like slice $\hat{x}^{\text{ct}}$ while preserving anatomy.

RAFM builds on Rectified Flow and formulates translation as deterministic transport from $\pi_{\text{cbct}}$ to $\pi_{\text{ct}}$. In practice, under small-data and small-batch medical training, random or batch-local pseudo pairing often yields semantically mismatched endpoint pairs and noisy transport targets. To address this, RAFM retrieves, for each CBCT slice, a feature-similar CT slice from a global CT memory bank in a frozen feature space, thereby constructing a more consistent empirical coupling (Fig.~\ref{fig:method}).

\subsection{Rectified Flow for Unpaired CBCT-to-CT Translation}

\subsubsection{The naive RF for image translation.}
Rectified Flow (RF) learns a time-dependent velocity field $v_\theta(x,t)$ and defines a deterministic transport ODE
\begin{equation}
\frac{d x_t}{d t} = v_\theta(x_t, t), \quad t\in[0,1].
\end{equation}
In our setting, the transport starts from a CBCT sample and ends at a CT sample: $x_0=x^{\text{cbct}}\sim\pi_{\text{cbct}}$ and $x_1=x^{\text{ct}}\sim\pi_{\text{ct}}$. The scalar $t\in[0,1]$ denotes the \emph{transport time}, where $t=0$ corresponds to the source endpoint $x_0$ and $t=1$ corresponds to the target endpoint $x_1$. RF constructs a linear interpolation path between endpoints and uses the intermediate state
\begin{equation}
x_t=(1-t)x_0+t x_1,\quad t\sim\mathcal{U}(0,1),
\label{eq:linear_interp}
\end{equation}
as the input to the velocity network at time $t$. Along this straight path, the target instantaneous velocity is constant, $u_t=x_1-x_0$, and RF trains $v_\theta$ by regressing to this target:
\begin{equation}
\mathcal{L}_{\text{RF}}(\theta)=
\mathbb{E}_{(x_0,x_1)\sim\rho,\; t\sim\mathcal{U}(0,1)}
\left[
\left\| v_\theta(x_t,t) - (x_1-x_0) \right\|_2^2
\right],
\end{equation}
where $\rho$ denotes an endpoint coupling used to sample $(x_0,x_1)$. Importantly, $\rho$ does not need voxel-aligned CBCT--CT correspondences; it suffices that its marginals match the two data distributions~\cite{rectified_flow}:
\begin{equation}
\rho_{x_0}=\pi_{\text{cbct}},\qquad \rho_{x_1}=\pi_{\text{ct}} .
\label{eq:distribution}
\end{equation}
Under this condition, RF learns the conditional mean velocity induced by $\rho$ at each $(x_t,t)$, providing a fully non-adversarial way to transport $\pi_{\text{cbct}}$ to $\pi_{\text{ct}}$ via ODE integration at inference~\cite{flow_matching,rectified_flow}.

\subsubsection{Advantage and limitation of RF.}
For CBCT-to-CT translation, the goal is to correct CBCT-specific artifacts and HU bias while keeping patient anatomy unchanged. Rectified flow learns a deterministic ODE along the straight interpolation in Eq.~\eqref{eq:linear_interp}, a formulation that has been repeatedly adopted for image-to-image translation where the output preserves the source content/structure while changing appearance or attributes~\cite{rectified_flow,flowie,flow_matching}. In this sense, RF is \emph{theoretically} compatible with the unpaired setting: the RF objective only requires an endpoint coupling $\rho$ with correct marginals (Eq.~\eqref{eq:distribution}),
rather than voxel-aligned paired correspondence. In practice, however, medical datasets are small and high-resolution training often forces very small mini-batches, so the empirical coupling induced by naive random pairing or batch-local matching can be a poor finite-sample approximation of the desired distribution-level coupling, leading to semantically mismatched endpoints and noisy transport targets~\cite{FM_pairing,FM_batch_OT}. This motivates Sec.~\ref{sec:RAFM}: we improve the practical coupling quality under small-data, small-batch conditions by introducing a global retrieval-based pseudo-pair construction strategy, which better realizes RF's straight-line transport for anatomy-sensitive unpaired translation.

\subsection{Retrieval-Augmented Flow Matching and Inference}\label{sec:RAFM}
\subsubsection{CT slice memory bank construction.}
While standard RF permits arbitrary couplings with correct marginals, RAFM improves practical training by using global retrieval to construct a more reliable empirical coupling under limited medical data. We use a frozen DINOv3~\cite{dinov3} encoder $f(\cdot)$ to extract a class-token embedding $z=f(x)\in\mathbb{R}^d$ for each slice and maintain a fixed-capacity CT memory bank
\begin{equation}
\mathcal{M} = \left\{(z_j^{\text{ct}}, x_j^{\text{ct}})\right\}_{j=1}^{K},
\end{equation}
where $z_j^{\text{ct}}$ is the DINOv3 feature for CT slice $x_j^{\text{ct}}$. Although the CT dataset is fixed, full-repository retrieval is inefficient in slice-wise training and may offer limited gains. Moreover, precomputing a single set of global features is less general under data augmentation, where slice appearance (and thus features) may vary across iterations. To balance efficiency and coupling quality, we implement $\mathcal{M}$ as a \emph{rolling} (online FIFO) \emph{training-global} buffer that aggregates CT slices across iterations. This bank is larger than a mini-batch but smaller than the full dataset, serving as a practical approximation to dataset-level retrieval. Concretely, each iteration enqueues CT feature--slice pairs from the current mini-batch into $\mathcal{M}$ (FIFO when $|\mathcal{M}|>K$), and retrieves feature-similar CT slices for the CBCT mini-batch from the updated bank.

\subsubsection{Retrieval-Augmented Flow Matching.}
For each CBCT slice $x_i^{\text{cbct}}$ in the current mini-batch, we compute $z_i^{\text{cbct}}=f(x_i^{\text{cbct}})$ and retrieve the most similar CT slice from $\mathcal{M}$ by cosine similarity, i.e., $j^*=\arg\max_j \mathrm{cos}(z_i^{\text{cbct}}, z_j^{\text{ct}})$. We then construct a pseudo endpoint pair $(x_0,x_1)=(x_i^{\text{cbct}}, x_{j^*}^{\text{ct}})$. The retrieved CT slice is selected from the target-domain memory bank solely by feature similarity, without using subject identity, temporal correspondence, registration, or paired CBCT--CT annotations. Therefore, RAFM remains a strictly unpaired  framework. Feature similarity does not imply voxel-wise anatomical alignment, so the retrieved CT slice is used as an endpoint sample for coupling construction rather than a direct regression target.

The retrieval procedure induces a retrieval-based empirical coupling $\rho_{\text{retr}}$, which replaces random pairing or batch-local matching during training. Under this coupling, the RAFM objective is
\begin{equation}
\mathcal{L}_{\text{RAFM}}(\theta)=
\mathbb{E}_{(x_0,x_1)\sim \rho_{\text{retr}},\; t\sim \mathcal{U}(0,1)}
\left[
\left\|v_\theta\big((1-t)x_0+t x_1,\; t\big)-(x_1-x_0)\right\|_2^2
\right].
\end{equation}
Compared with random or batch-wise pseudo pairing, retrieval-augmented pairing provides more semantically consistent transport supervision by expanding the candidate pool from the mini-batch to a global CT pool. We parameterize $v_\theta(x,t)$ using a time-conditioned U-Net~\cite{songunet}. At inference, given $x(0)=x^{\text{cbct}}$, we solve the learned ODE forward from $t=0$ to $t=1$ and take the final state as the synthesized CT slice $\hat{x}^{\text{ct}}$.
\begin{algorithm}[t]
\caption{Training of Retrieval-Augmented Flow Matching (RAFM)}
\label{alg:rafm}
\begin{algorithmic}[1]
\STATE \textbf{Input:} true-unpaired training sets $\mathcal{D}^{\text{tr}}_{\text{cbct}}=\{x_i^{\text{cbct}}\}_{i\in\mathcal{A}}$, $\mathcal{D}^{\text{tr}}_{\text{ct}}=\{x_i^{\text{ct}}\}_{i\in\mathcal{B}}$, where $\mathcal{A}\cap\mathcal{B}=\varnothing$, memory-bank size $K$
\STATE \textbf{Initialize:} velocity network $v_\theta$, frozen encoder $f$, empty CT memory queue $\mathcal{M}$
\FOR{each training iteration}
    \STATE Sample a CBCT mini-batch $\{x_i^{\text{cbct}}\}_{i=1}^{B}$ from $\mathcal{D}^{\text{tr}}_{\text{cbct}}$
    \STATE Sample a CT mini-batch $\{x_m^{\text{ct}}\}_{m=1}^{B'}$ from $\mathcal{D}^{\text{tr}}_{\text{ct}}$
    \STATE Compute CT features $z_m^{\text{ct}}=f(x_m^{\text{ct}})$ for all $m$
    \STATE Enqueue $\{(z_m^{\text{ct}},x_m^{\text{ct}})\}_{m=1}^{B'}$ into $\mathcal{M}$; if $|\mathcal{M}|>K$, discard the oldest entries (FIFO)
    \STATE Initialize mini-batch loss $\mathcal{L}\leftarrow 0$
    \FOR{each $x_i^{\text{cbct}}$}
        \STATE Compute $z_i^{\text{cbct}}=f(x_i^{\text{cbct}})$
        \STATE Retrieve $x_{j^*}^{\text{ct}}$ from $\mathcal{M}$ by cosine similarity:
        \STATE \hspace{1em} $j^*=\arg\max_j \ \mathrm{cos}(z_i^{\text{cbct}}, z_j^{\text{ct}})$
        \STATE Form pseudo pair $(x_0,x_1)\leftarrow (x_i^{\text{cbct}}, x_{j^*}^{\text{ct}})$
        \STATE Sample $t\sim\mathcal{U}(0,1)$ and form $x_t=(1-t)x_0+t x_1$
        \STATE $\mathcal{L}\leftarrow \mathcal{L}+\|v_\theta(x_t,t)-(x_1-x_0)\|_2^2$
    \ENDFOR
    \STATE $\mathcal{L}\leftarrow \mathcal{L}/B$
    \STATE Update $\theta$ using $\mathcal{L}$
\ENDFOR
\end{algorithmic}
\end{algorithm}

\section{Experiments}

\subsection{Experimental Setup}

We evaluate RAFM on the SynthRAD2023 CBCT-to-CT benchmark (pelvis)~\cite{SynthRAD2023}, which contains 180 paired 3D cases. The dataset is split at the subject level into training/validation/test sets with a 7:1:2 ratio (126/18/36). To construct a strict true-unpaired setting, the 126 training subjects are further divided into two disjoint groups: CBCT volumes from 63 subjects and CT volumes from the remaining 63. Thus, no CBCT--CT correspondence or cross-modal subject overlap is available during training, while paired volumes are retained only for validation and testing. This subject-level protocol is stricter than slice-level shuffling.

We adopt slice-wise 2D training by decomposing each 3D volume into slices. All slices are resized to $256\times256$, clipped to $[-1024,2000]$, and normalized to $[-1,1]$. The velocity field is parameterized by a time-conditioned U-Net with sinusoidal timestep embeddings and an MLP projection~\cite{songunet}. For RAFM, we use a frozen DINOv3-base encoder~\cite{dinov3} for slice-level feature extraction and cosine-similarity retrieval in a CT memory bank (default $K=512$, top-1). The model is trained with Adam (learning rate $2\times10^{-4}$, batch size 4) for 100 epochs. At inference, the learned ODE is solved using 10-step Euler integration.

For quantitative evaluation of the synthesized CT images, we report Mean Absolute Error (MAE), Structural Similarity (SSIM)~\cite{ms-ssim&psnr}, Peak Signal-to-Noise Ratio (PSNR)~\cite{ms-ssim&psnr}, Fr\'echet Inception Distance (FID)~\cite{fid}, and SegScore. SegScore is defined as the average Dice over pelvic organs segmented by TotalSegmentator~\cite{totalsegmentator} on synthesized and reference CT volumes. Unless otherwise specified, MAE/SSIM/PSNR/SegScore are computed on reconstructed 3D volumes, while FID is computed on 2D slices.

\subsection{Results}

We compare RAFM with representative unpaired CBCT-to-CT translation methods, including GAN-based methods (CycleGAN~\cite{cbct2ct_cycleGAN}, GcGAN~\cite{GcGAN}, CUT~\cite{CuT}) and diffusion/SB methods (SynDiff~\cite{syndiff}, UNSB~\cite{UNSB}), under the same subject-level true-unpaired protocol.

\begin{table*}[t]
\caption{Quantitative results on SynthRAD2023 for unpaired CBCT-to-CT translation. Best results are in bold. \dag denotes a significant improvement (p-value < 0.05) over existing method.}
\label{tab:main_results}
\centering
\small
\setlength{\tabcolsep}{3.5pt}
\renewcommand{\arraystretch}{0.95}
\begin{tabular}{lccccc}
\toprule
Method & MAE (HU)$\downarrow$ & SSIM (\%)$\uparrow$ & PSNR (dB)$\uparrow$ & FID$\downarrow$ & SegScore (\%)$\uparrow$ \\
\midrule
CycleGAN~\cite{cbct2ct_cycleGAN} & 173.8$\pm$14.2 & 77.96$\pm$0.78 & 23.57$\pm$1.62 & 99.75 & 64.03$\pm$1.98 \\
GcGAN~\cite{GcGAN}              & 212.9$\pm$18.6 & 76.38$\pm$0.86 & 22.71$\pm$1.74 & 102.6 & 59.34$\pm$2.14 \\
CUT~\cite{CuT}                  & 109.0$\pm$12.1 & 80.27$\pm$0.62 & 24.47$\pm$1.28 & 66.77 & 70.49$\pm$1.58 \\
SynDiff~\cite{syndiff}          & 104.2$\pm$11.5 & 77.68$\pm$0.74 & 24.94$\pm$1.21 & 72.36 & 62.58$\pm$1.87 \\
UNSB~\cite{UNSB}                & 110.8$\pm$12.8 & 80.03$\pm$0.65 & 24.42$\pm$1.33 & 62.91 & 72.07$\pm$1.46 \\
RAFM (ours)                     & \textbf{101.2$\pm$10.4}$^\dag$ & \textbf{80.96$\pm$0.49}$^\dag$ & \textbf{25.15$\pm$1.12}$^\dag$ & \textbf{53.29}$^\dag$ & \textbf{75.77$\pm$1.21}$^\dag$ \\
\bottomrule
\end{tabular}
\end{table*}

\begin{figure*}[t]
\centering
\includegraphics[width=\textwidth]{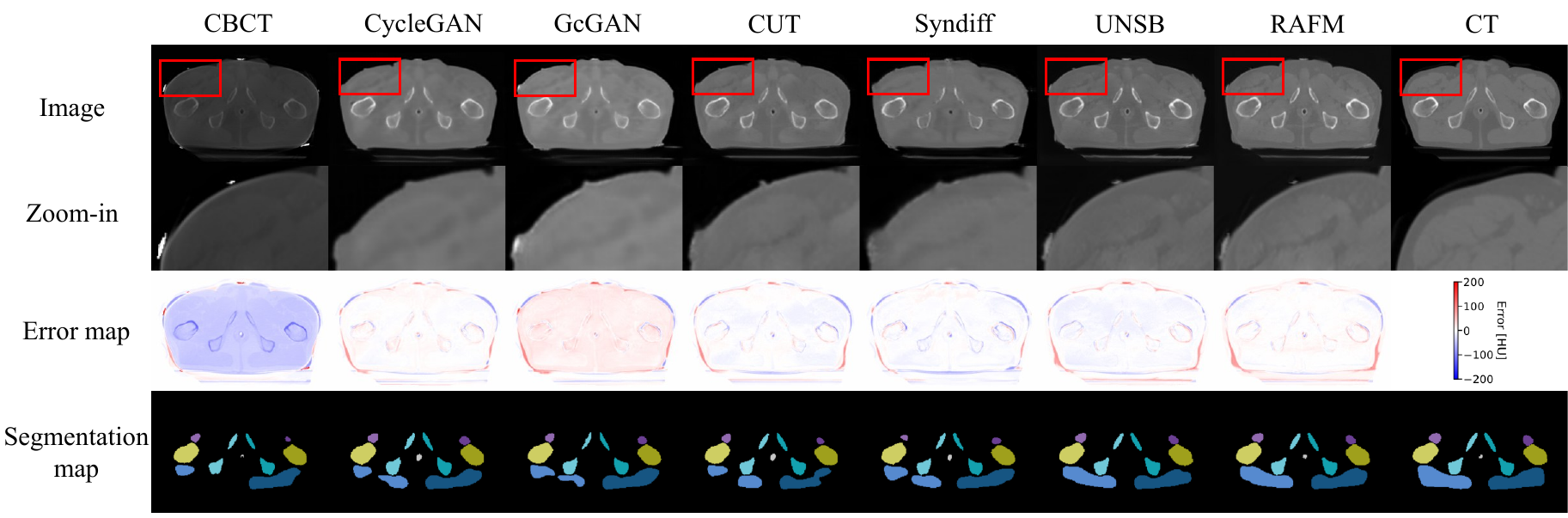}
\caption{Qualitative comparison on SynthRAD2023 for unpaired CBCT-to-CT translation.}
\label{fig:qualitative}
\end{figure*}

As shown in Table~\ref{tab:main_results}, RAFM consistently outperforms prior unpaired methods under the subject-level true-unpaired protocol. Compared to the strongest baselines, RAFM reduces MAE to $101.2$ (vs.\ $104.2$ for SynDiff and $110.8$ for UNSB) and achieves the highest SSIM/PSNR ($80.96\%$ and $25.15$). Notably, RAFM yields the best distributional realism with the lowest FID (53.29), improving over UNSB (62.91) and CUT (66.77). It also improves anatomy-related consistency, reaching a SegScore of $75.77\%$, significantly outperforming UNSB ($72.07\%$) and CUT ($70.49\%$). Qualitative examples in Fig.~\ref{fig:qualitative} further show cleaner artifact suppression and more stable anatomical structures achieved by our method than the others. 
In terms of computational cost, RAFM trains a single forward CBCT$\rightarrow$CT transport model, with training overhead comparable to a standard time-conditioned U-Net, avoiding the doubled generators and cycle-consistency constraints required by bidirectional frameworks such as CycleGAN~\cite{cbct2ct_cycleGAN} and SynDiff~\cite{syndiff}. At inference, RAFM uses only 10 ODE steps, making it substantially faster than multi-step diffusion sampling, while remaining within a practical runtime range compared to GAN-based baselines.

\subsection{Ablation Study}

We conduct a unified ablation to study coupling quality and candidate-pool size in true-unpaired training. We compare: (i) a conventional unpaired regression baseline (vanilla U-Net~\cite{Unet} without flow matching), (ii) RF with random unpaired coupling, (iii) RF with batch-wise matching, and (iv) RAFM with retrieval-augmented coupling using different memory-bank sizes. For a unified view, random coupling is treated as the degenerate case with no retrieval pool ($K=0$), and batch-wise matching is treated as a local strategy with candidate pool size equal to the mini-batch size ($K=4$). We also report a paired RF model trained with full paired CBCT--CT supervision as a supervised reference (upper bound).

\begin{table*}[t]
\caption{Unified ablation on coupling strategy and candidate-pool size $K$ on SynthRAD2023. $K{=}0$ denotes random coupling and $K{=}4$ denotes batch-wise matching; paired RF is shown as a reference (upper bound).}
\label{tab:ablation_unified}
\centering
\resizebox{\textwidth}{!}{
\begin{tabular}{lc|ccccc}
\toprule
Training strategy & $K$ & MAE (HU)$\downarrow$ & SSIM (\%)$\uparrow$ & PSNR (dB)$\uparrow$ & FID$\downarrow$ & SegScore (\%)$\uparrow$ \\
\midrule
Vanilla U-Net (w/o FM)      & 0    & 127.1  & 67.70 & 23.01 & 209.8 & 1.45 \\
RF + random coupling        & 0    & 132.2  & 77.20 & 23.76 & 96.15 & 74.27 \\
RF + batch-wise matching    & 4    & 118.4  & 79.84 & 24.41 & 71.32 & 75.01 \\
\midrule
RAFM                        & 256  & 109.2  & 80.46 & 24.75 & 53.74 & 75.34 \\
RAFM                        & 512  & 101.2 & \textbf{80.96} & \textbf{25.15} & \textbf{53.29} & \textbf{75.77} \\
RAFM                        & 1024 & \textbf{98.50}  & 80.68 & \textbf{25.15} & 55.41 & 75.29 \\
RAFM                        & 4096 & 103.0  & 80.81 & 25.11 & 58.06 & 74.94 \\
\midrule
Paired RF (upper bound)     & /    & 72.28  & 86.30 & 27.12 & 59.78 & 76.87 \\
\bottomrule
\end{tabular}
}
\end{table*}

Table~\ref{tab:ablation_unified} shows a clear performance improvement as coupling quality and candidate-pool size increase. The vanilla U-Net baseline (w/o FM) fails under true-unpaired supervision (FID $209.8$, SegScore $1.45\%$), confirming that direct regression to arbitrary targets is ineffective. In contrast, RF remains robust even with random coupling (FID $96.15$, SegScore $74.27\%$), and batch-wise matching ($K{=}4$) further improves MAE and FID. By enlarging the candidate pool beyond the mini-batch, RAFM consistently outperforms local matching and achieves the best true-unpaired trade-off at $K{=}512$ (MAE $101.2$, FID $53.29$, SegScore $75.77\%$), with diminishing returns for larger banks. Paired RF provides an upper bound on voxel-wise fidelity (MAE $72.28$, SSIM $86.30\%$), while RAFM remains close in anatomy-related metrics (SegScore $75.77\%$ vs.\ $76.87\%$), indicating that semantically meaningful coupling enables stable anatomy preservation even without paired supervision.

\section{Conclusion}

We proposed Retrieval-Augmented Flow Matching (RAFM), a non-adversarial rectified-flow framework for unpaired CBCT-to-CT translation with retrieval-augmented coupling. Using a frozen DINOv3 encoder and a global CT memory queue, RAFM constructs more reliable pseudo pairs than random or batch-local matching under small-data, small-batch training. Experiments on SynthRAD2023 with a strict subject-level true-unpaired protocol demonstrate strong image quality and anatomy consistency, and ablations highlight coupling quality as a key factor for narrowing the gap to paired rectified-flow training.

%
%
%
\bibliographystyle{splncs04}
\bibliography{refS}

@article{cbct_application_review,
  title={Guidelines for clinical use of {CBCT}: a review},
  author={Horner, Keith and O'Malley, Lucy and Taylor, Kathryn and Glenny, Anne-Marie},
  journal={Dentomaxillofacial radiology},
  volume={44},
  number={1},
  pages={20140225},
  year={2015},
  publisher={Oxford University Press}
}

@article{sct_review,
  title={Deep learning based synthetic-{CT} generation in radiotherapy and {PET}: a review},
  author={Spadea, Maria Francesca and Maspero, Matteo and Zaffino, Paolo and Seco, Joao},
  journal={Medical physics},
  volume={48},
  number={11},
  pages={6537--6566},
  year={2021},
  publisher={Wiley Online Library}
}

@article{cbct_artifacts,
  title={Artefacts in {CBCT}: a review},
  author={Schulze, Ralf and Heil, Ulrich and Gro$\beta$, D and Bruellmann, Dan Dominik and Dranischnikow, Egor and Schwanecke, Ulrich and Schoemer, Elmar},
  journal={Dentomaxillofacial Radiology},
  volume={40},
  number={5},
  pages={265--273},
  year={2011},
  publisher={Oxford University Press}
}

@article{cbct2ct_cycleGAN,
  title={{CBCT}-based synthetic {CT} generation using deep-attention cycleGAN for pancreatic adaptive radiotherapy},
  author={Liu, Yingzi and Lei, Yang and Wang, Tonghe and Fu, Yabo and Tang, Xiangyang and Curran, Walter J and Liu, Tian and Patel, Pretesh and Yang, Xiaofeng},
  journal={Medical physics},
  volume={47},
  number={6},
  pages={2472--2483},
  year={2020},
  publisher={Wiley Online Library}
}

@inproceedings{Unet,
  title={{U-Net}: Convolutional networks for biomedical image segmentation},
  author={Ronneberger, Olaf and Fischer, Philipp and Brox, Thomas},
  booktitle={MICCAI},
  pages={234--241},
  year={2015}
}

@article{SynthRAD2023,
  title={SynthRAD2023 Grand Challenge dataset: Generating synthetic {CT} for radiotherapy},
  author={Thummerer, Adrian and van der Bijl, Erik and Galapon Jr, Arthur and Verhoeff, Joost JC and Langendijk, Johannes A and Both, Stefan and van den Berg, Cornelis (Nico) AT and Maspero, Matteo},
  journal={Medical physics},
  volume={50},
  number={7},
  pages={4664--4674},
  year={2023},
  publisher={Wiley Online Library}
}

@inproceedings{GcGAN,
  title={Geometry-consistent generative adversarial networks for one-sided unsupervised domain mapping},
  author={Fu, Huan and Gong, Mingming and Wang, Chaohui and Batmanghelich, Kayhan and Zhang, Kun and Tao, Dacheng},
  booktitle={Proceedings of the IEEE/CVF conference on computer vision and pattern recognition},
  pages={2427--2436},
  year={2019}
}

@inproceedings{CuT,
  title={Contrastive learning for unpaired image-to-image translation},
  author={Park, Taesung and Efros, Alexei A and Zhang, Richard and Zhu, Jun-Yan},
  booktitle={European conference on computer vision},
  pages={319--345},
  year={2020},
  organization={Springer}
}

@article{syndiff,
  title={Unsupervised medical image translation with adversarial diffusion models},
  author={{\"O}zbey, Muzaffer and Dalmaz, Onat and Dar, Salman UH and Bedel, Hasan A and {\"O}zturk, {\c{S}}aban and G{\"u}ng{\"o}r, Alper and Cukur, Tolga},
  journal={IEEE Transactions on Medical Imaging},
  volume={42},
  number={12},
  pages={3524--3539},
  year={2023},
  publisher={IEEE}
}

@inproceedings{UNSB,
  title={UNPAIRED IMAGE-TO-IMAGE TRANSLATION VIA NEURAL SCHR{\"O}DINGER BRIDGE},
  author={Kim, Beomsu and Kwon, Gihyun and Kim, Kwanyoung and Ye, Jong Chul},
  booktitle={12th International Conference on Learning Representations, ICLR 2024},
  year={2024}
}

@inproceedings{ACSB,
  title={Anatomy-Conserving Unpaired CBCT-to-CT Translation via Schr{\"o}dinger Bridge},
  author={Shi, Ke and Ouyang, Song and Liu, Gang and Luo, Yong and Su, Kehua and Liang, Zhiwen and Du, Bo},
  booktitle={International Conference on Medical Image Computing and Computer-Assisted Intervention},
  pages={46--55},
  year={2025},
  organization={Springer}
}

@article{flow_matching,
  title={Flow matching for generative modeling},
  author={Lipman, Yaron and Chen, Ricky TQ and Ben-Hamu, Heli and Nickel, Maximilian and Le, Matt},
  journal={arXiv preprint arXiv:2210.02747},
  year={2022}
}

@article{rectified_flow,
  title={Flow straight and fast: Learning to generate and transfer data with rectified flow},
  author={Liu, Xingchao and Gong, Chengyue and Liu, Qiang},
  journal={arXiv preprint arXiv:2209.03003},
  year={2022}
}

@inproceedings{FM_batch_OT,
  title={Contrastive flow matching},
  author={Stoica, George and Ramanujan, Vivek and Fan, Xiang and Farhadi, Ali and Krishna, Ranjay and Hoffman, Judy},
  booktitle={Proceedings of the IEEE/CVF International Conference on Computer Vision},
  pages={1185--1194},
  year={2025}
}

@article{FM_pairing,
  title={Gradient Variance Reveals Failure Modes in Flow-Based Generative Models},
  author={Reu, Teodora and Dromigny, Sixtine and Bronstein, Michael and Vargas, Francisco},
  journal={arXiv preprint arXiv:2510.18118},
  year={2025}
}

@article{dinov3,
  title={Dinov3},
  author={Sim{\'e}oni, et al.,Oriane},
  journal={arXiv preprint arXiv:2508.10104},
  year={2025}
}

@article{songunet,
  title={Score-based generative modeling through stochastic differential equations},
  author={Song, Yang and Sohl-Dickstein, Jascha and Kingma, Diederik P and Kumar, Abhishek and Ermon, Stefano and Poole, Ben},
  journal={arXiv preprint arXiv:2011.13456},
  year={2020}
}

@article{ms-ssim&psnr,
  title={A multimodal comparison of latent denoising diffusion probabilistic models and generative adversarial networks for medical image synthesis},
  author={M{\"u}ller-Franzes, et al.,Gustav},
  journal={Scientific Reports},
  volume={13},
  number={1},
  pages={12098},
  year={2023},
  publisher={Nature Publishing Group UK London}
}

@article{totalsegmentator,
  title={TotalSegmentator: robust segmentation of 104 anatomic structures in {CT} images},
  author={Wasserthal, et al.,Jakob},
  journal={Radiology: Artificial Intelligence},
  volume={5},
  number={5},
  pages={e230024},
  year={2023},
  publisher={Radiological Society of North America}
}

@article{fid,
  title={Gans trained by a two time-scale update rule converge to a local nash equilibrium},
  author={Heusel, Martin and Ramsauer, Hubert and Unterthiner, Thomas and Nessler, Bernhard and Hochreiter, Sepp},
  journal={Advances in neural information processing systems},
  volume={30},
  year={2017}
}

@inproceedings{flowie,
  title={Flowie: Efficient image enhancement via rectified flow},
  author={Zhu, Yixuan and Zhao, Wenliang and Li, Ao and Tang, Yansong and Zhou, Jie and Lu, Jiwen},
  booktitle={Proceedings of the IEEE/CVF Conference on Computer Vision and Pattern Recognition},
  pages={13--22},
  year={2024}
}

\end{document}